\documentclass{article}

\usepackage[english]{babel}
\usepackage{multirow} 
\usepackage[letterpaper,top=2cm,bottom=2cm,left=3cm,right=3cm,marginparwidth=1.75cm]{geometry}

\usepackage{amsmath}
\usepackage{pifont}
\usepackage{graphicx}

\usepackage[x11names]{xcolor}
\usepackage[most]{tcolorbox}
\usepackage[colorlinks = true,
            linkcolor = blue,
            urlcolor  = blue,
            citecolor = blue,
            anchorcolor = blue]{hyperref}
\newcommand{\xmark}{\ding{55}}
\newcommand{\cmark}{\ding{51}}

\newtcolorbox{shadedcvbox}[1][]{enhanced jigsaw,
  colback=white!80!blue,
  coltext={black},
  boxrule=0pt,
  arc=3mm,
  auto outer arc,
  boxsep=3pt,
  left=4pt,
  right=2pt,
  bottom=2pt,
  top=2pt,
  fontupper={\bfseries},
  #1}

\definecolor{codegreen}{rgb}{0,0.6,0}
\definecolor{codegray}{rgb}{0.5,0.5,0.5}
\definecolor{codepurple}{rgb}{0.58,0,0.82}
\definecolor{backcolour}{rgb}{0.95,0.95,0.92}

\lstdefinestyle{mystyle}{
  backgroundcolor=\color{backcolour}, commentstyle=\color{codegray},
  keywordstyle=\color{blue},
  numberstyle=\tiny\color{codegray},
  stringstyle=\color{codepurple},
  basicstyle=\ttfamily\footnotesize,
  breakatwhitespace=false,         
  breaklines=true,                 
  captionpos=b,                    
  keepspaces=true,                 
  numbers=left,                    
  numbersep=5pt,                  
  showspaces=false,                
  showstringspaces=false,
  showtabs=false,                  
  tabsize=2
}

\title{Improving Segment Anything on the Fly: Auxiliary Online Learning and Adaptive Fusion for Medical Image Segmentation}
\author{Tianyu Huang, Tao Zhou, Weidi Xie, Shuo Wang, Qi Dou, Yizhe Zhang\thanks{T.Huang, T. Zhou and Y. Zhang are with the School of Computer
Science and Engineering, Nanjing University of Science and Technology,
Nanjing, China. S. Wang is with the Digital Medical Research Center, School of Basic Medical Sciences, Fudan University, and Shanghai Key Laboratory of MICCAI, Shanghai, China. W. Xie is with Shanghai Jiaotong University, Shanghai, China. Q. Dou is with the Department of Computer Science and Engineering, the Chinese University of Hong Kong, Hong Kong. \protect\\
E-mail: yizhe.zhang.cs@gmail.com}
}
\date{}

\begin{document}
\maketitle

\begin{center}
  Project Link: \href{https://sam-auxol.github.io/AuxOL/}{https://sam-auxol.github.io/AuxOL/}
\end{center}

\begin{abstract}
The current variants of the Segment Anything Model (SAM), which include the original SAM and Medical SAM, still lack the capability to produce sufficiently accurate segmentation for medical images. In medical imaging contexts, it is not uncommon for human experts to rectify segmentations of specific test samples 
after SAM generates its segmentation predictions. These rectifications typically entail manual or semi-manual corrections employing state-of-the-art annotation tools. Motivated by this process, we introduce a novel approach that leverages the advantages of online machine learning to enhance Segment Anything (SA) during test time. We employ rectified annotations to perform online learning, with the aim of improving the segmentation quality of SA on medical images. To improve the effectiveness and efficiency of online learning when integrated with large-scale vision models like SAM, we propose a new method called Auxiliary Online Learning (AuxOL). AuxOL creates and applies a small auxiliary model (specialist) in conjunction with SAM (generalist), entails adaptive online-batch and adaptive segmentation fusion. Experiments conducted on eight datasets covering four medical imaging modalities validate the effectiveness of the proposed method. Our work proposes and validates a new, practical, and effective approach for enhancing SA on downstream segmentation tasks (e.g., medical image segmentation). 
\end{abstract}

\section{Introduction}

\begin{quote} 
\centering 
``It is all about how \textit{effectively} and \textit{efficiently} we can utilize the Segment Anything Model (SAM) for downstream tasks.''
\end{quote}

The Segment Anything Model (SAM)~\cite{kirillov2023segment}, since its introduction, has drawn tremendous attention and effort in the field of medical image analysis. Image segmentation plays a critical role in medical AI systems, and there is a great potential in applying large vision models such as SAM to medical images. Pioneering medical AI studies have proposed (1) training a medical SAM~\cite{ma2024segment,cheng2023sam}, or (2) fine-tuning and/or adapting SAM for downstream medical image segmentation tasks~\cite{wu2023medical,zhang2023customized}. Training a medical SAM requires acquiring a large volume of labeled medical images. There exist practical challenges in building SAM for medical image data, including variations in modalities (e.g., CT, MRI, Ultrasound), subjects (e.g., organs, tissues, cells), scales, annotation quality, and tasks. On the other hand, fine-tuning and/or adapting a general-purpose SAM to a particular medical image segmentation task, although a reasonable approach, can lead to the possibility of SAM losing its generalization capability (to certain degree). For each distinct medical image segmentation task, one may need to apply full-session offline fine-tuning and/or adaptation. Although possible to do, it incurs significant computation and time costs, and being rigid as each task requires a new session of offline training.


\begin{figure}[t]
\centering
\includegraphics[width=1.0\textwidth]{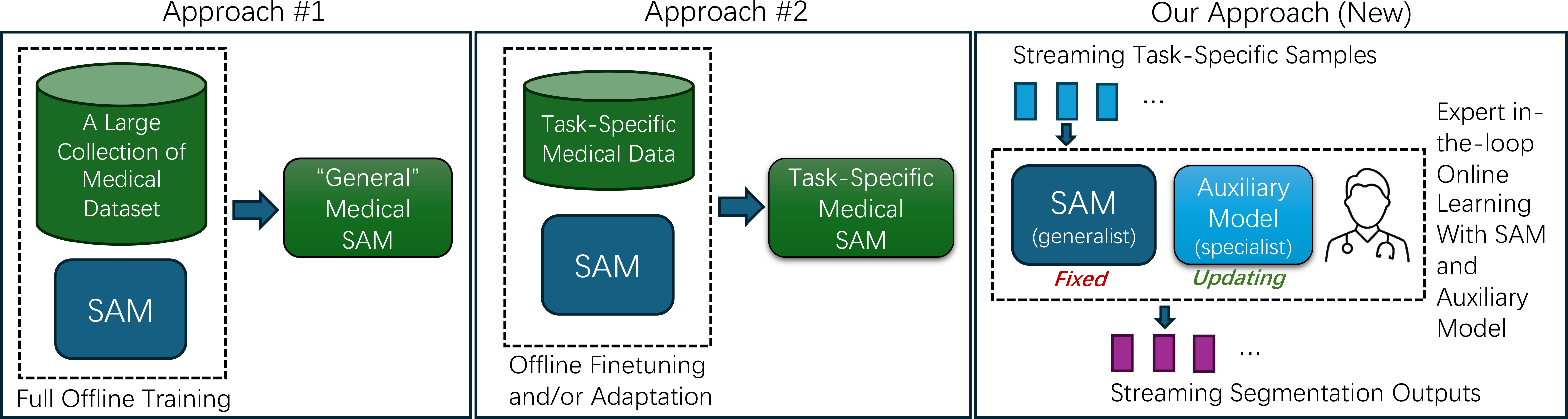}
\caption{Existing and our approaches on utilizing SAM for downstream medical image segmentation.} \label{fig:approach}
\end{figure}

In this paper, we propose a new approach for improving Segment Anything (SA) on medical images, leveraging the advantages of online machine learning (OML). OML is a common scenario when human expert feedback is available during deployment of a machine learning model. One can utilize such feedback to improve the model's performance on the fly in test time (without large-scale offline retraining and/or fine-tuning). To improve the effectiveness, efficiency and stability of online learning with SAM, we propose a new method that utilizes a much smaller (compared to SAM) auxiliary model during inference in conjunction with SAM. The auxiliary model adjusts SAM's output while conducting online weight updates. We call this online learning method AuxOL (Auxiliary Online Learning). AuxOL is applicable to the original SAM~\cite{kirillov2023segment}, newly developed medical SAM~\cite{ma2024segment}, and already adapted SAM (e.g., with inserted adaption layers~\cite{wu2023medical}). The unique approach-level characteristics of our proposed method are depicted in Figure \ref{fig:approach}, which provides an overview of the method's distinctiveness.
Our main contributions can be summarized in four-fold.
\begin{itemize}
\item To our best knowledge, for the first time, we propose to leverage the advantages and utilities of OML to improve SAM on medical image segmentation. 

\item We develop a new method called AuxOL, which creates and applies a small auxiliary model (specialist) in conjunction with SAM (generalist) during deployment and online learning, improving learning effectiveness, efficiency and stability.

\item Comprehensive experiments on eight segmentation datasets in four medical imaging modalities demonstrate the advantages of combining OML with SAM and the effectiveness of the proposed AuxOL method.

\item Capitalizing on human expertise and the adaptability of online learning, our method presents a new promising avenue for advancing SAM-based medical image segmentation.

\end{itemize}






\section{Related Work}

\subsection{Medical Segment Anything Model}
Ma et al.~\cite{ma2024segment} collected 84 public datasets consisting of over one million images with 10 imaging modalities, and trained a Segment Anything Model for medical images. In a similar fashion, Cheng et al.~\cite{cheng2023sam} trained their medical SAMs, with two models released, one for 2D images and another for 3D images. Recently, Zhao et al.~\cite{zhao2023one} proposed a segment anything model using text prompts (SAT model). A large dataset was curated which includes over 11K 3D medical image scans from 31 segmentation datasets. Careful standardization on both visual scans and label space has been conducted when constructing the dataset. Text query is used together with the input image for prompting SAT in generating segmentation masks corresponding to the content specified in the text. However, there still exist moderate to large performance gaps between the current medical SAMs' performance and the level of segmentation performance deemed acceptable for clinical usage. Early studies have also indicated that the learned features in medical foundation models do not necessarily lead to significant performance improvements when compared to much smaller known models~\cite{alfasly2023foundation}. \textcolor{black}{Effectively and efficiently utilizing foundation models for downstream medical imaging tasks undoubtedly offers significant benefits but also presents substantial challenges.}


\subsection{Adapting SAM for Specific Downstream Tasks}
Medical-SAM-Adapter~\cite{wu2023medical} proposed to insert additional layers to SAM for adapting SAM to downstream medical image segmentation tasks. During adaptation, only the inserted layers are updated using labeled samples. SAMed~\cite{zhang2023customized} used LoRA~\cite{hu2021lora} for SAM adaptation; instead of adding new layers between the existing layers in SAM, it adds layers in the $q$ and $v$ projection layers of each Transformer block in the image encoder along with the existing layers. MA-SAM~\cite{chen2023masam} added a 3D adapter to the Transformer block of the image encoder so that it can be used on various volumetric or video medical data, effectively extracting 3D information with 2D SAM pre-trained weights. SAMUS~\cite{lin2023samus} introduced a parallel CNN branch based on SAM and developed a position adapter and a feature adapter to adapt SAM from natural images to medical images. SAC~\cite{na2024segment} proposed a new nucleus prompt generation method and used LoRA technology similar to SAMed.  \textcolor{black}{Existing SAM adaptation methods rely on offline full-training approaches, which incur significant computation and time costs. Additionally, they are rigid, as each new task requires a new session of offline training.}

\subsection{Online Machine Learning}
Online machine learning (OML) has long been a prevalent strategy for numerous real-world applications, praised for its flexibility and efficiency. A comprehensive survey of online learning can be found in~\cite{hoi2021online}. Here, we briefly describe online learning techniques related to our work. Zinkevich~\cite{zinkevich2003online} proposed online gradient descent (OGD) for the online learning setup, which can be viewed as an online version of the gradient descent algorithm and has been widely applied in online learning. Sahoo et al.~\cite{sahoo2017online} proposed the Hedge Back-propagation (HBP) algorithm, which addresses challenges by learning models of adaptive depths from a sequence of training data in an OML setting. Zhao et al.~\cite{2014Online} proposed Online Transfer Learning. Applying Online Meta-Learning (OML) to large vision foundation models, such as SAM, is intriguing. \textcolor{black}{However, the substantial sizes of modern foundation models like SAM present significant challenges in terms of costs, effectiveness, and stability.}

\begin{figure}[t]
\centering
\includegraphics[width=1.0\textwidth]{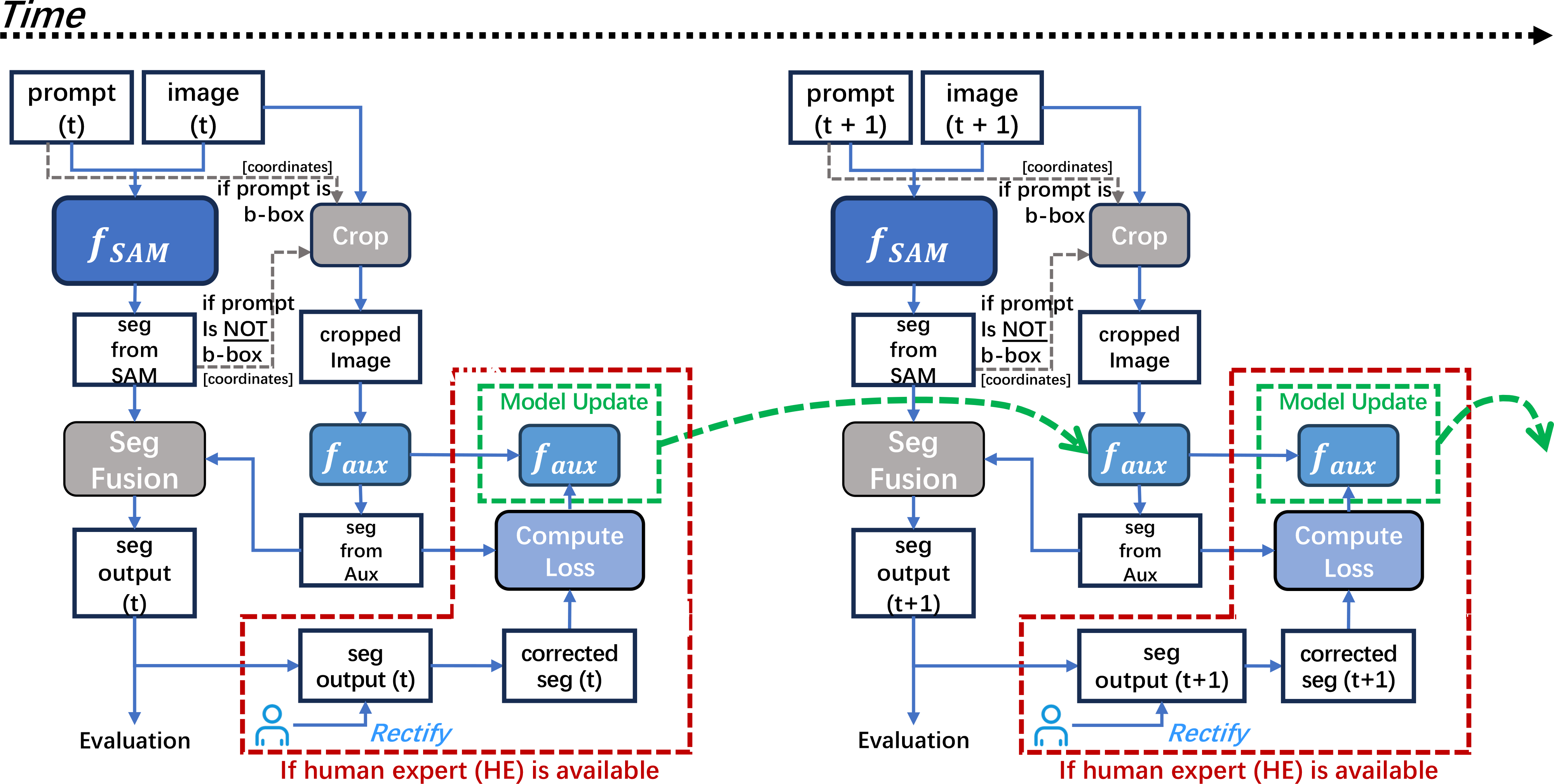}
\caption{An overview of the main steps of our AuxOL with SAM: Improving Segment Anything (SA) for medical images via auxiliary learning in an online learning pipeline.} \label{fig:overview}
\end{figure}

\section{Auxiliary Online Learning with SAM}

\subsection{Problem Setup}

Test samples are sequentially provided to the segmentation system. Specifically, at time $t$, a test image sample $x_t$ is provided to the segmentation system, and at time $t+1$, another sample $x_{t+1}$ is provided. 
Given a Segment Anything Model $f_{SAM}$, we aim to apply $f_{SAM}$ to the current image sample $x_t$ to obtain segmentation masks according to the particular medical image segmentation task at hand. Since SAM and its variants require points and/or bounding boxes as prompts for generating segmentation masks, we assume that each object of interest in $x_t$ is given at least a prompt (e.g., a center point or a bounding box). This problem setup was utilized in previous work (e.g., \cite{ma2024segment,huang2024segment}).




\subsection{Main Steps}\label{sec:ms}
An overview of our proposed auxiliary online learning method, AuxOL with SAM, is given in Figure~\ref{fig:overview}. We create and utilize a small auxiliary model (denoted as $f_{aux}$), which works along with SAM for model inference and online learning. The architecture of this auxiliary model can be as simple as a classic U-Net~\cite{ronneberger2015u}. The auxiliary model can be initialized with random weights. It is worth mentioning that when ImageNet pre-trained encoder weights are available, one can start with a pre-trained encoder and a randomly initialized decoder.

During model inference, a test image sample $x_t$ and a set of prompts ($P_t = \{p_1^t, p_2^t, \dots,p_{m_t}^t\}$) are given, where $m_t$ denotes the number of prompts for the image $x_t$. For every prompt, we feed $x_t$ and $p^t_j$, $j = 1, 2, \dots, m_t$, to $f_{SAM}$, and obtain a segmentation prediction $s_{t,j}$ of SAM. For each $j$, we obtain a cropped region $x_{t,j}$ based on the coordinates extracted from the prompt. When a bounding-box (b-box) is used as prompt, we directly use the b-box to crop the image region inside that b-box. When a center point is used as prompt, we compute the convex hull of the segmented areas in $s_{t,j}$, and then compute the b-box of the convex hull for image cropping. Next, we provide the cropped region $x_{t,j}$ to $f_{aux}$, and obtain segmentation output from this auxiliary model, denoted as $u_{t,j}$. Note that both $x_{t,j}$ and $u_{t,j}$ are in the logits form. We then proceed to fuse $s_{t,j}$ and $u_{t,j}$ to generate the segmentation output $\hat{y}_{t,j}$. The segmentation fusion function can be designed in many different ways. 


For simplicity, we combine these two maps using a scalar parameter $\alpha$ to balance the contributions from the two models, as:
\begin{equation}
\label{eq1}
\hat{y}_{t,j} =  \tau (\alpha s_{t,j} + (1-\alpha) u_{t,j}),
\end{equation}
where $\tau$ is an activation function, e.g., Sigmoid function. By default, $\alpha$ is set to 0.5. In Sec.~\ref{sec:ao}, we further develop a method to automatically configure the value of $\alpha$.

After the inference step is conducted for the sample $x_t$ and the prompt $p^t_j$, \textbf{if a human expert (HE) is available}, our method asks HE to look at the segmentation $\hat{y}_{t,j}$ and give manual correction on $\hat{y}_{t,j}$ if needed (possibly using semi-automatic annotation tools). The rectification generates a corrected segmentation map, denoted as $y_{t,j}$, which can be used to perform online learning to update the auxiliary model. Such update could be helpful for future test samples as the auxiliary model is actively learning the new sample data distribution. If HE is not available, one can consider using techniques such as pseudo-labeling~\cite{lee2013pseudo} and self-supervised learning~\cite{ouyang2022self} to generate loss values for the auxiliary model update. In this section, we focus on the supervised setting for method illustration.

Given the rectified segmentation map $y_{t,j}$, suppose a loss function is pre-defined as $L(\cdot)$ (e.g., Dice loss). A simple way to compute the online learning loss is:
\begin{equation}
loss_{t,j} = L(f_{aux}(x_{t,j}), y_{t,j}).
\end{equation}

One can use back-propagation (BP) to update the auxiliary model using the loss thus computed. A more sophisticated way to perform model update is to use online-batch based BP. This requires constructing and maintaining an online-batch as we process through new test samples. A straightforward way to do so is to store previous test samples with their rectified segmentations, and add the most recent sample to the online-batch for computing the batch-loss and executing BP. Suppose we set the online-batch size as $k$ (e.g., $k$ = 32) at the early time of online learning. When the number of processed samples (with rectified segmentations) is $< k$, we simply add the new sample to the online-batch without additional treatment. However, with more samples processed, the online-batch will increase in size, and it is not reasonable nor practical to keep an ever-increasing online-batch without pruning or dropping samples which the model already learned well. Hence, when a new sample is considered for the online-batch and the online-batch size is already $k$, we choose the sample with the smallest loss (obtained during the last time of the $f_{aux}$ update) in the online-batch, and compare this smallest loss with the loss obtained for the current sample. If the loss value of the current sample is higher, then we use the current sample to replace the sample with the smallest loss in the online-batch, keeping the batch size as $k$; otherwise, we keep the original online-batch unchanged. We then perform forward and backward propagations on the auxiliary model using the online-batch. The optimizer is set as AdamW~\cite{loshchilov2017decoupled} with the learning rate set as 0.0005. 

We give the pseudo-code of the main steps of AuxOL below.

\begin{lstlisting}[language=Python, caption= AuxOL: Online learning with the auxilary model and human feedback.]
def AuxOL(sam, aux, x, p, ol_batch):
    # sam: segment anything model, aux: auxiliary model.
    # x: image, p: prompt, ol_batch: contains previous samples.
    y_sam = sam(x, p) # inference on SAM using image and prompt.
    x_c = cropping(x,get_xywh(y_sam))
    #get_xywh is a function to retrieve bounding-box coordinates.
    u = aux(x_c) # inference on auxilary model.
    seg_output = seg_fusion(y_sam,u) #fusing the two maps.
    if human_expert is avaiable:
        y = human_expert_rectify(seg_output) #human-in-the-loop.
        loss = compute_loss(u,y) # loss of current sample.
        if ol_batch is not full:
            ol_batch.insert(x_c,y,loss) # insert the newest.            
        elif ol_batch.smallest_loss <= loss:
            ol_batch.remove_smallest() # remove the smallest.
            ol_batch.insert(x_c,y,loss) # insert the newest.
        u_batch = aux(ol_batch.x_c) # using all the samples.
        loss_batch = compute_loss(u_batch, ol_batch.y)
        ol_batch.update(loss_batch)
        aux <- backprop(aux, loss_batch) # update auxilary model.
    return seg_output, aux, ol_batch
\end{lstlisting}

\subsection{Advanced Options}\label{sec:ao}
\noindent\textbf{Adaptive Segmentation Fusion.} When combining/fusing the segmentation outputs from SAM and the auxiliary model, a simplest way is to use a fixed value of $\alpha$ in Eq.~(\ref{eq1}) (e.g., $\alpha$ = 0.5). However, to achieve a more optimal fusion effect, one can consider determining the value of $\alpha$ adaptively according to the statistics obtained from the previously processed samples. For each test sample, once the segmentation output is produced and HE performs corrections on the segmentation map, with the obtained rectified segmentation, we can conduct an optimization process to find an optimal fusion parameter for this sample. Given the segmentation from the SAM ($s_{t,j}$), the segmentation from the auxiliary model ($u_{t,j}$), and the rectified segmentation ($y_{t,j}$), we aim to minimize the following objective with respect to the value of $\alpha$, as:
\begin{equation}
 \arg\max_{\alpha_{t,j}} \mathcal{DSC}(\tau(\alpha_{t,j} s_{t,j} + (1-\alpha_{t,j}) u_{t,j}) , y_{t,j})\label{eq:optimal_alpha},
\end{equation}
where $\mathcal{DSC}$ represents the Dice similarity coefficient. We denote the optimal value of $\alpha_{t,j}$ thus obtained as $\alpha_{t,j}^*$. There are many options for optimizing the above objective. One can choose to use an iterative method such as gradient descent or an exhaustive method with a finite sampling of valid solutions in the space of $\alpha$. We record each found optimal $\alpha_{t,j}^*$. For the current test sample, we estimate its value by taking the mean of the optimal $\alpha^*$ values for the last $K$ inference instances. This mean is used to set the current ${\alpha}$ value for segmentation fusion. By default, we set $K$ to 5.

\noindent\textbf{SAM's Output as Part of the Input of the Auxiliary Model.}
The design described in Sec.~\ref{sec:ms} employs a typical segmentation model for constructing the auxiliary model, whose input is a cropped raw image region. Another possible approach is to provide both the segmentation output of SAM and the cropped raw image region to the auxiliary model, denoted as $u_{t,j} = f_{aux}(s_{t,j},x_{t,j})$. This approach treats the auxiliary model as a refinement model, refining the segmentation output of SAM rather than generating the segmentation solely based on the raw image region.

A potential risk with this design arises when SAM's output is of very low quality (e.g., if the segmentation map is nearly arbitrary). Feeding such low-quality segmentation to the auxiliary model can lead to sub-optimal results. Specifically, for point prompts, SAM's output quality is generally lower than for bounding-box prompts. Consequently, we feed SAM's output to the auxiliary model only when the prompt is a bounding-box, and provide only the raw image region to the auxiliary model when the prompt is a point.




\section{Experiments}

\subsection{AuxOL on SAM and Medical SAM}
We investigate the effectiveness of the proposed AuxOL method with SA for downstream medical image segmentation tasks. We utilize polyp segmentation in endoscopic images (5 datasets)~\cite{fan2020pranet}, gland segmentation in histology images (1 dataset)~\cite{sirinukunwattana2017gland}, breast cancer segmentation in ultrasound images (1 dataset)~\cite{al2020dataset}, and fluid region segmentation in OCT scans (1 dataset)~\cite{bogunovic2019retouch}. All these datasets are publicly accessible. In this subsection, we consider the setting in which, after inference, the segmentation prediction of every image sample is rectified by a human expert. We use ground truth masks provided by the test set of each dataset to serve as rectified segmentations.

The original SAM-H~\cite{kirillov2023segment} ({641.1M}) and Medical SAM~\cite{ma2024segment} (93.7M) are tested, and a U-Net~\cite{ronneberger2015u} (\textbf{17.3M}) that is initialized with random weights serves as the auxiliary model in our AuxOL. Table~\ref{tab:1} and Table~\ref{tab:2} show that AuxOL significantly improves the segmentation performances of SAM and Medical SAM in both the Dice score (DSC) and Hausdauff Distance (HD) metrics. 

Note that in clinical practice, one may choose to use rectified segmentation maps provided by human experts for clinical applications. In our context, we employ segmentation outputs from the segmentation models and compare them with ground truths, i.e., the rectified segmentation maps provided by human experts, in order to evaluate the performances of SAM and Medical SAM, along with assessing the effect of AuxOL. Additionally, one can perceive AuxOL as a method to alleviate the burden of manual rectification by human experts, as the segmentation maps are significantly enhanced by AuxOL, hence reducing the number of pixel locations that require rectification by human experts.

\begin{table}[t]
    \centering
    \scriptsize
    \caption{Online learning of AuxOL improves the performances of SAM and Medical SAM in polyp segmentation of endoscopic images (five datasets).}
    \begin{tabular}{ |c| c | c| c| c | c | c | c | c|}  
        \hline  
         Model & Prompt & \textbf{AuxOL} &Metrics & ClinicDB~\cite{bernal2015wm} & ColonDB~\cite{tajbakhsh2015automated} & ETIS~\cite{silva2014toward} & Kvasir~\cite{jha2020kvasir} & CVC-300~\cite{bernal2012towards} \\  
        \hline

        \multirow{8}{*}{SAM-H~\cite{kirillov2023segment}} & \multirow{4}{*}{B-Box}& \multirow{2}{*}{\xmark}& DSC $\uparrow$& 92.34 & 89.88& 92.63 &90.34& 93.04 \\ 
         \cline{4-9}
                                                                     & & & HD$\downarrow$ &\textbf{15.04} & 27.16 & 32.69 & \textbf{39.79} & 14.84\\  
        \cline {3-9}
        &  & \multirow{2}{*}{\cmark} &DSC$\uparrow$& \textbf{92.63}&\textbf{92.73}	&\textbf{94.21}&\textbf{92.75}	&\textbf{95.43}\\  
         \cline{4-9}
                                                                               &&&HD$\downarrow$ &15.97&\textbf{20.48}&	\textbf{25.10}&	61.83&	\textbf{8.37}\\ 
          \cline {2-9}
          &\multirow{4}{*}{Point} &\multirow{2}{*}{\xmark}&DSC$\uparrow$& 62.98& 60.23 & 56.73 & 73.87 & 76.49 \\  
           \cline{4-9}
                                                                       & & &HD$\downarrow$ &80.77 & 138.07 & 275.64 & 97.64 & 69.24\\ 
         \cline{3-9}
           &&\multirow{2}{*}{\cmark} &DSC$\uparrow$& \textbf{63.36}&	\textbf{68.10}&	\textbf{62.70}&	\textbf{75.93}&	\textbf{78.14}\\  
            \cline{4-9}
                                                          &  & &HD$\downarrow$ &\textbf{79.44}	&\textbf{108.48}&	\textbf{219.16}	&\textbf{94.52}&	\textbf{68.60}\\ 
         \hline
         \hline
          \multirow{8}{*}{MedSAM~\cite{ma2024segment}} &\multirow{4}{*}{B-Box} &\multirow{2}{*}{\xmark} &DSC$\uparrow$& 80.65 & 76.59& 81.62& 87.39 & 79.11 \\  
           \cline{4-9}
                                                                           &  & &HD$\downarrow$ &23.26 & 40.00 & 45.22 & \textbf{63.71} & 23.20\\ 
         \cline{3-9}
          & &\multirow{2}{*}{\cmark} &DSC$\uparrow$& \textbf{85.36}	&\textbf{90.59}	&\textbf{91.62}	&\textbf{92.11}	&\textbf{93.74}\\  
           \cline{4-9}
                                                                            &  & &HD$\downarrow$  &\textbf{21.26}	&\textbf{24.95}	&\textbf{35.52}	&65.79	&\textbf{11.35}\\ 
         \cline{2-9}
           &\multirow{4}{*}{Point}&\multirow{2}{*}{\xmark} &DSC$\uparrow$& 25.99 &21.37 & 32.35 & 42.56 & 10.32 \\  
           \cline{4-9}
                                                                           &  & &HD$\downarrow$ &173.57 & 301.36 & 610.74 & 274.96 & 351.36\\ 
        \cline{3-9}
           &&\multirow{2}{*}{\cmark}&DSC$\uparrow$& \textbf{31.92}	&\textbf{59.18}&\textbf{41.32}&\textbf{58.58}&\textbf{72.34}\\  
            \cline{4-9}
                                                                           &  & &HD$\downarrow$ &\textbf{154.96}	&\textbf{148.69}	&\textbf{442.32}	&\textbf{192.91}	&\textbf{113.85}\\ 
         \hline
    
    \end{tabular} 
    \label{tab:1}
\end{table}

\begin{table}[t]
    \centering
    \scriptsize
    \caption{Online learning of AuxOL improves the performances of SAM and Medical SAM in breast cancer segmentation of ultrasound images, gland segmentation of histology images, and fluid region segmentation of OCT scans.}
    \begin{tabular}{ |c|c|c|c| c c |c  c| c c c|   }  
    \hline  
     \multirow{2}{*}{ Model }& \multirow{2}{*}{Prompt}& \multirow{2}{*}{\textbf{AuxOL}}& \multirow{2}{*}{Metrics} &\multicolumn{2}{c|}{BUSI~\cite{al2020dataset}} &\multicolumn{2}{c|}{GlaS~\cite{sirinukunwattana2017gland}}  &\multicolumn{3}{c|}{Fluidchallenge~\cite{bogunovic2019retouch}}\\
      \cline{5-11}
    &&&& Benign & Malignant & Benign & Malignant   & Cirrus & Topcon & Spectralis  \\  
    \hline  
    
    \multirow{8}{*}{SAM-H~\cite{kirillov2023segment}} &\multirow{4}{*}{B-Box}& \multirow{2}{*}{\xmark}&DSC$\uparrow$ & 86.45 & 83.47 &89.55& 84.47 & 82.13& 85.90 & 87.19  \\  
     \cline{4-11}
                                                                       &&&HD$\downarrow$ &27.71 & 57.70 & \textbf{74.36} & 103.96&68.13 & 55.05 & 43.67  \\ 
     
     \cline{3-11}
     
    && \multirow{2}{*}{\cmark}&DSC$\uparrow$& \textbf{92.98}	&\textbf{87.03}&\textbf{91.02}	&\textbf{86.13}&\textbf{95.63}&\textbf{96.79}&\textbf{96.34}\\  
      \cline{4-11}
                                                                    && &HD$\downarrow$ &\textbf{15.1}&	\textbf{47.74}&79.69	&\textbf{101.47}&	\textbf{25.01}&\textbf{	17.62}&\textbf{	14.56}\\ 
     \cline{2-11}
     &\multirow{4}{*}{Point}& \multirow{2}{*}{\xmark} &DSC$\uparrow$& 66.59& 57.97 & 53.72 & 61.14 & 59.14 & 58.38& 56.18 \\  
       \cline{4-11}
                                                                   && &HD$\downarrow$ &96.77& 154.68& 143.80 & 181.04&206.32 & 185.29 & 175.59  \\ 
     \cline{3-11}
    & &\multirow{2}{*}{\cmark}&DSC$\uparrow$&\textbf{72.41}	&\textbf{67.29}	&\textbf{79.26}&	\textbf{76.56}&	\textbf{83.50}&	\textbf{85.74}	&\textbf{75.99}  \\  
      \cline{4-11}
                                                                     &&&HD$\downarrow$ &\textbf{87.86}&\textbf{116.32}&	\textbf{125.02}&	\textbf{138.81}&	\textbf{88.13}&	\textbf{68.85}	&\textbf{102.35} \\ 
     \hline
     \hline
      \multirow{8}{*}{MedSAM~\cite{ma2024segment}} &\multirow{4}{*}{B-Box} &\multirow{2}{*}{\xmark} &DSC$\uparrow$& 82.74 & 80.91 & 85.86& 88.42 & 74.75& 63.91 & 56.01  \\  
        \cline{4-11}
                                                      &  & &HD$\downarrow$ &26.41 & 52.43 & 84.70 & 104.21 &80.33&77.37&	59.91\\ 
     \cline{3-11}
       &&\multirow{2}{*}{\cmark}  &DSC$\uparrow$& \textbf{92.41}&	\textbf{84.34}&	\textbf{89.95}	&\textbf{88.50}	&\textbf{94.91}&	\textbf{95.67}&	\textbf{94.49}  \\  
         \cline{4-11}
                                                                         &  & &HD$\downarrow$ &\textbf{16.00}	&\textbf{48.70}&	\textbf{81.67}&	\textbf{99.93}	&\textbf{32.67}&	\textbf{22.67}&	\textbf{25.61}\\ 
     \cline{2-11}
       &\multirow{4}{*}{Point} &\multirow{2}{*}{\xmark} &DSC$\uparrow$& 42.07&45.54& 60.73 & 64.44 & 38.17& 28.59& 27.21 \\  
         \cline{4-11}
                                                                       &  & &HD$\downarrow$ &239.04&194.54& 186.93 & 187.49 & 378.14 &311.11& 152.61\\ 
     \cline{3-11}
    &&\multirow{2}{*}{\cmark}  &DSC$\uparrow$& \textbf{65.41}&	\textbf{62.61}	&\textbf{83.71}&	\textbf{83.33}	&\textbf{89.68}	&\textbf{90.24}	&\textbf{74.05}\\ 
      \cline{4-11}
                                                              &  & &HD$\downarrow$ &\textbf{126.37}&\textbf{145.26}	&\textbf{143.64}&	\textbf{165.75}	&\textbf{70.47}	&\textbf{59.71}	&\textbf{105.01} \\ 
     \hline
\end{tabular}   
\label{tab:2}
\end{table}


\subsection{AuxOL with Partial Supply of Human Expert Feedback}

\begin{figure}[t]
    \centering
    \includegraphics[width=1.0\linewidth]{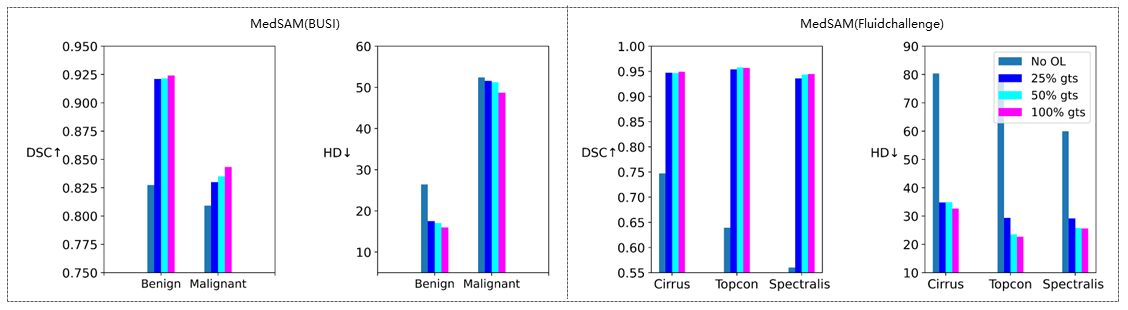}
    \caption{AuxOL under varying levels of ground truth supply.}
    \label{fig:fig2}
\end{figure}

In this subsection, we evaluate the performance of AuxOL when human experts rectify only a portion of test samples. Figure~\ref{fig:fig2} presents the cases when 0\%, 25\%, 50\%, and 100\% of samples receive rectified segmentations (ground truth) under a uniform sampling rate. We observe that a bigger amount of human expert feedback corresponds to better segmentation improvement. Remarkably, even when only 25\% of samples receive segmentation rectification and are engaged in the online learning process, AuxOL still notably enhances the segmentation results.

\begin{figure}[t]
    \centering
    \includegraphics[width=1.0\linewidth]{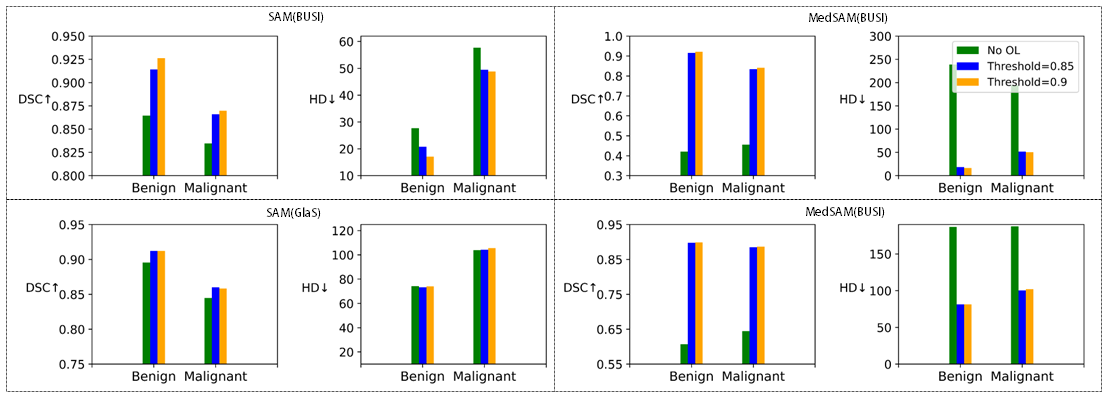}
    \caption{AuxOL with varying thresholds (in Dice score) to provoke HE rectifications.}
    \label{fig:threshold}
\end{figure}

In addition, we delve deeper into a more practical scenario in which human experts correct only those samples exhibiting low-quality segmentation. We simulate this situation by initially comparing ground truth images with segmentation outputs using the DSC evaluation metric. When the resulting Dice score falls below a certain threshold (e.g., 0.85), we supply ground truth images (HE-rectified segmentations) to facilitate AuxOL. As Figure~\ref{fig:threshold} shows, it is evident that in such situations, AuxOL can offer support to SAM segmentation, especially in the cases when only a limited amount of feedback from human experts is available.

\subsection{AuxOL on Already-Adapted SAM (Task-Specific)}

\begin{figure}[!t]
    \centering
    \includegraphics[width=1.0\linewidth]{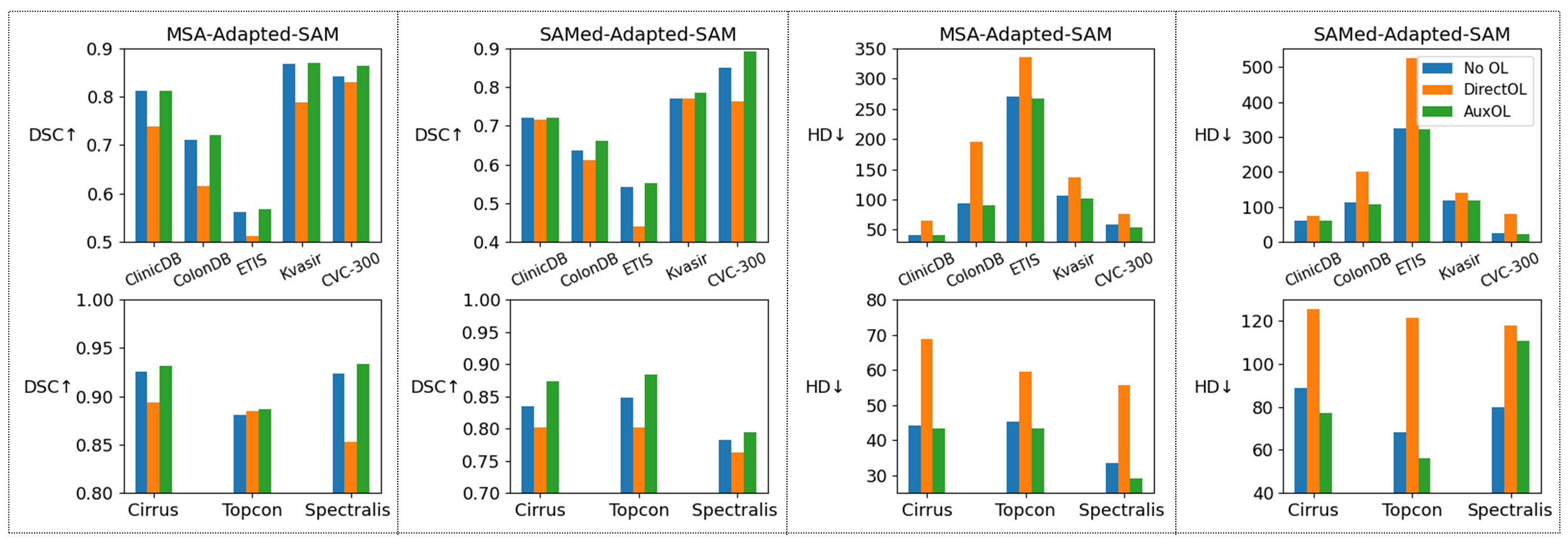}
    \caption{Online learning performance of AuxOL with Already-Adapted SAM.}
    \label{fig:fig4}
\end{figure}

We also examine the effect of AuxOL in the context when SAM has already been adapted to a downstream task (e.g., polyp segmentation). An adaptation method utilizes a training set of the downstream task for adapting SAM to that task. In our experiments, we employ two state-of-the-art SAM adaptation methods: MSA~\cite{wu2023medical} and SAMed~\cite{zhang2023customized}. Both MSA and SAMed insert additional trainable layers into SAM during training-time adaptation. After the adaptation is complete, we apply AuxOL during test time to test samples for online learning with the already-adapted SAM. From Figure~\ref{fig:fig4}, we observe that AuxOL effectively improves already-adapted SAM in most cases. Moreover, one can opt to directly update the added adaptation layers (by MSA or SAMed) in SAM during online learning (called the DirectOL approach~\cite{zhou2017machine}). We compare our AuxOL with DirectOL, and find that AuxOL outperforms DirectOL. DirectOL can degrade segmentation performance in some cases, since directly updating model parameters in an online fashion may suffer from instability issues and lead to a model collapse problem.

\begin{table}[!]
    \centering
    \scriptsize
    \caption{Effectiveness of using online-batch based model updates in AuxOL.}
    \begin{tabular}{ |c|c|c| c c c  c c| }  
    \hline  
      \multirow{2}{*}{Model}& \multirow{2}{*}{Options}& \multirow{2}{*}{Metrics} &\multicolumn{5}{c|}{Polyp}\\
       \cline{4-8}
    &&& ClinicDB & ColonDB & ETIS & Kvasir & CVC-300\\

  \hline
    \multirow{4}{*}{\shortstack{SAM-H~\cite{kirillov2023segment}}}&\multirow{2}{*}{Single Sample}& DSC$\uparrow$&92.61&	91.49&	93.58&	91.41&	94.50\\  
    \cline{3-8}
     & &HD$\downarrow$ &16.02	&22.81	&28.94	&63.18	&11.82\\ 
    \cline{2-8}
     &\multirow{2}{*}{\textbf{Online-Batch}}& DSC$\uparrow$& \textbf{92.63}&\textbf{92.73}&	\textbf{94.21}&	\textbf{92.75}&\textbf{95.43}\\  
     \cline{3-8}
     & &HD$\downarrow$ &\textbf{15.97}&	\textbf{20.48}	&\textbf{25.1}&	\textbf{61.83}&	\textbf{8.37}\\ 
        \hline
         \hline
    \multirow{4}{*}{MedSAM~\cite{ma2024segment}}&\multirow{2}{*}{Single Sample}& DSC$\uparrow$& 82.53&85.23&	87.39	&90.94	&92.65\\ 
    \cline{3-8}
     & &HD$\downarrow$ &22.26&35.32&	40.93&	67.66&	14.19\\ 
    \cline{2-8}
     &\multirow{2}{*}{\textbf{Online-Batch}}& DSC$\uparrow$& \textbf{85.36}	&\textbf{90.59}&	\textbf{91.62}	&\textbf{92.11}	&\textbf{93.74}\\  
     \cline{3-8}
     & &HD$\downarrow$ &\textbf{21.26}&\textbf{24.95}&\textbf{35.52}	&\textbf{65.79}	&\textbf{11.35}\\ 
    \hline
    \end{tabular}
    \label{tab:ab1}
\end{table}

   
  

\begin{table}[!]
    \centering
    \scriptsize
    \caption{Effectiveness of using online-batch based model updates in AuxOL.}
    \begin{tabular}{ |c|c|c| c c | c c| c c c| }  
    \hline  
      \multirow{2}{*}{Model}& \multirow{2}{*}{Options}& \multirow{2}{*}{Metrics} &\multicolumn{2}{c|}{BUSI}&\multicolumn{2}{c|}{GlaS}&\multicolumn{3}{c|}{Fluidchallenge}\\
      \cline{4-10}
     &&& Benign & Malignant & Benign & Malignant   & Cirrus & Topcon & Spectralis \\

   \hline
    \multirow{4}{*}{SAM-H~\cite{kirillov2023segment}}&\multirow{2}{*}{Single Sample}& DSC$\uparrow$&91.39&85.79&90.74	&85.45&92.94&	95.50	&94.70\\  
    \cline{3-10}
     & &HD$\downarrow$ &20.35&51.07&78.49	&103.93&40.41	&26.46	&21.59\\ 
    \cline{2-10}
     &\multirow{2}{*}{\textbf{Online-Batch}}& DSC$\uparrow$& \textbf{92.98}&	\textbf{87.03}&\textbf{91.02}	&\textbf{86.13}& \textbf{95.63}	&\textbf{96.79}	&\textbf{96.34}\\  
     \cline{3-10}
                              & &HD$\downarrow$ &\textbf{15.1}	&\textbf{47.74 }&\textbf{79.69}&\textbf{101.47} &\textbf{25.01}&\textbf{17.62}	&\textbf{14.56}\\
      \hline
       \hline
    \multirow{4}{*}{MedSAM~\cite{ma2024segment}}&\multirow{2}{*}{Single Sample}& DSC$\uparrow$&91.00&82.39&88.55	&88.48&92.14	&92.85&92.24\\  
    \cline{3-10}
     & &HD$\downarrow$ &18.36&50.98 &97.27&104.52&57.65&43.06&35.26\\ 
    \cline{2-10}
     &\multirow{2}{*}{\textbf{Online-Batch}}& DSC$\uparrow$& \textbf{92.41}&\textbf{84.34 }&\textbf{89.95}&\textbf{88.50}&\textbf{94.91}&\textbf{95.67}&\textbf{94.49}\\  
     \cline{3-10}
     & &HD$\downarrow$ &\textbf{16.00}&\textbf{48.70} &\textbf{81.67}&\textbf{99.93} &\textbf{32.67}&\textbf{22.67}&\textbf{25.61}\\ 
    \hline
    \end{tabular}
    \label{tab:ab}
\end{table}

\begin{table}[!]
    \centering
    \scriptsize
    \caption{Effectiveness of the adaptive segmentation fusion.}
    \begin{tabular}{ |c|c|c| c c| c c c| }  
   
    \hline
   \multirow{2}{*}{Model}& \multirow{2}{*}{Options}& \multirow{2}{*}{Metrics} &\multicolumn{2}{c|}{GlaS}&\multicolumn{3}{c|}{Fluidchallenge}\\
    \cline{4-8}
    &&& Benign & Malignant   & Cirrus & Topcon & Spectralis \\  
   
    \hline
    \multirow{4}{*}{SAM-H~\cite{kirillov2023segment}}&\multirow{2}{*}{Fixed (0.5)}& DSC$\uparrow$&67.69&	69.69&75.36&	75.67&	66.94\\ 
    \cline{3-8}
     & &HD$\downarrow$ &126.05	&154.3&159.75&	140.75&	158.5\\ 
    \cline{2-8}
     &\multirow{2}{*}{\textbf{Adaptive}}& DSC$\uparrow$ &\textbf{79.26}	&\textbf{76.56}&\textbf{83.50}&\textbf{85.74}&	\textbf{75.99}\\  
     \cline{3-8}
     & &HD$\downarrow$  &\textbf{125.02}	&\textbf{138.81}&\textbf{88.13}&	\textbf{68.85}&	\textbf{102.35}\\ 
        \hline 
           \hline
    \multirow{4}{*}{MedSAM~\cite{ma2024segment}}&\multirow{2}{*}{Fixed (0.5)}& DSC$\uparrow$&83.02	&81.93&82.16	&78.13	&65.42\\ 
    \cline{3-8}
     & &HD$\downarrow$&145.08	&166.12&143.23	&162.87&	\textbf{103.76}\\ 
    \cline{2-8}
     &\multirow{2}{*}{\textbf{Adaptive}}& DSC$\uparrow$&\textbf{83.71}&	\textbf{83.33}& \textbf{89.68}	&\textbf{90.24}	&\textbf{74.05}\\ 
     \cline{3-8}
     & &HD$\downarrow$  &\textbf{143.64}	&\textbf{165.75} &\textbf{70.47}	&\textbf{59.71}&	105.01\\

    \hline
    \end{tabular}
    \label{tab:ab_adativefusion}
\end{table}

\begin{figure}[!t]
    \centering
    \includegraphics[width=1.0\linewidth]{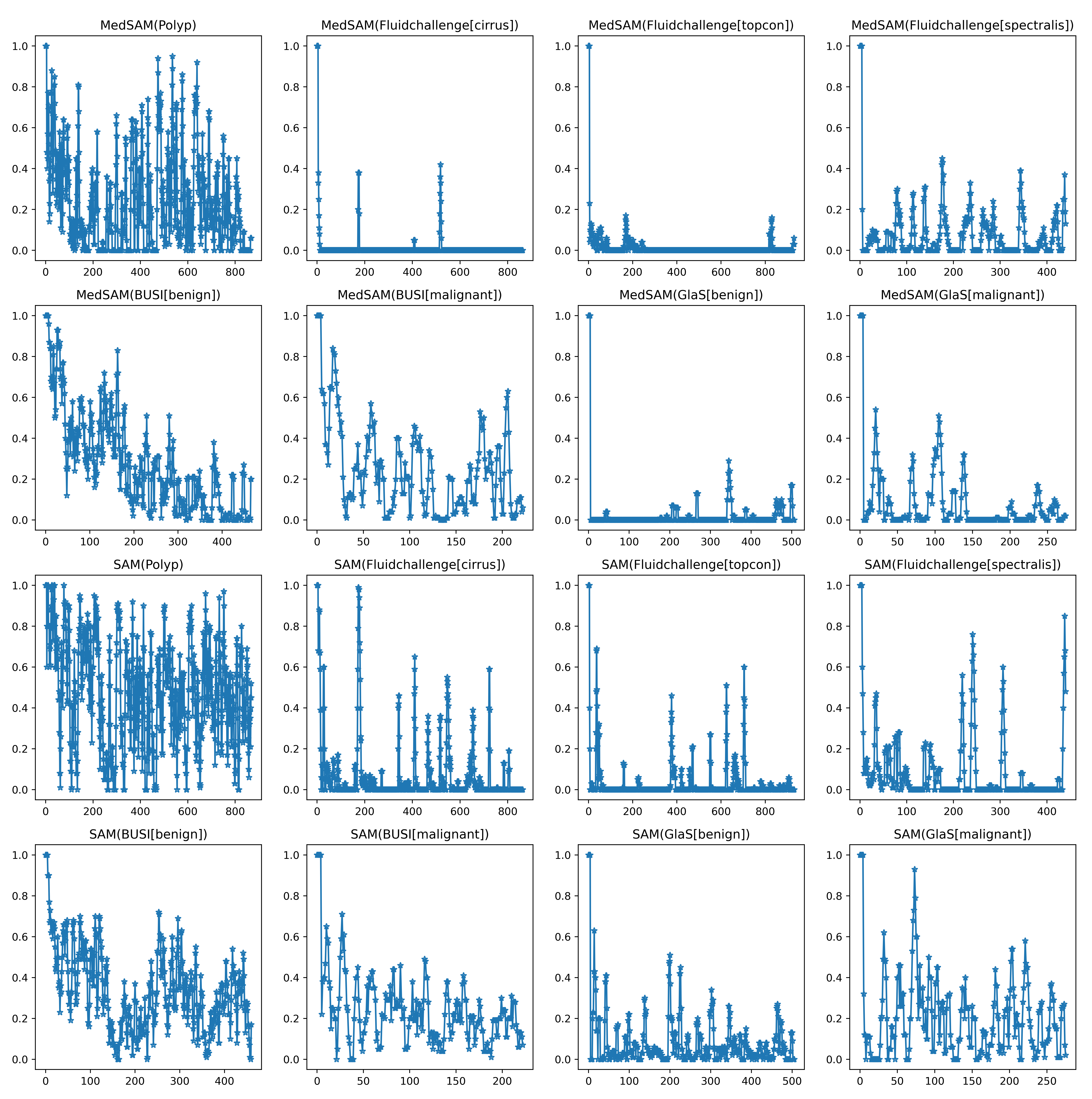}
    \caption{The value of $\alpha$ (used in adaptive fusion) over time. Each data point corresponds to an 
    image sample with a particular prompt in an inference sample sequence.}
    \label{fig:alpha}
    
\end{figure}

\subsection{Ablation and Additional Studies}

\textbf{Online-Batch.} We investigate the effectiveness of utilizing online-batch based model updates, and compare it with the single-sample based counterpart. From Table~\ref{tab:ab1} and Table~\ref{tab:ab}, we find that online-batch based model updates exhibit greater effectiveness. 

\textbf{Adaptive Fusion.} Table~\ref{tab:ab_adativefusion} shows that our adaptive fusion scheme significantly improves the accuracy of online learning with SAM and Medical SAM. To gain insight into the dynamics of segmentation fusion and the roles of SAM and the auxiliary model in this online learning framework, we present in Figure~\ref{fig:alpha} the evolving values of $\alpha$ in adaptive fusion over time. 
Note that $\alpha$ is estimated by computing the mean of the \textbf{optimal} $\alpha^*$ values for the previous $K$ instances of the current inference instance (see Sec.~\ref{sec:ao}). Thus, examining how $\alpha$ changes is useful in studying the roles of SAM and the auxiliary model in this online learning framework over time. Our observations are as follows. (1) In the initial stage, SAM plays a more crucial role in producing segmentation results (as $\alpha$ being closer to 1). (2) As the auxiliary model learns from samples, it effectively gathers task-specific segmentation knowledge, progressively contributing more to segmentation generation (indicated by $\alpha$ approaching 0). (3) Periodically, we observe an increase in the value of $\alpha$, indicating cases where SAM gives better segmentation than that of the auxiliary model, showing SAM's utility within this framework.





\section{Conclusions}

The practice of human experts rectifying AI-generated segmentation predictions is not uncommon in medical AI applications. This paper introduced a novel and practical method for enhancing the Segment Anything Model (SAM) for medical image segmentation. Rather than investing resources in time-consuming and energy-intensive offline fine-tuning, adaptation, or retraining of SAM for medical images, we advocated leveraging valuable feedback from human experts and harnessing the benefits of online learning for test-time improvement of Segment Anything (SA) in medical imaging scenes. Our proposed Auxiliary Online Learning (AuxOL) method is both effective and efficient. It eliminates the need for extensive tuning of SAM, offering potential utility across various downstream segmentation tasks. By capitalizing on human expertise and the adaptability of online learning, our method presents a promising avenue for advancing SAM-based medical image segmentation.

\bibliographystyle{plain}
\bibliography{ref}

\end{document}